\documentclass[journal,twoside,web]{ieeecolor}
\usepackage{generic}
\usepackage{cite}
\usepackage{amsmath,amssymb,amsfonts,mathtools}
\usepackage{algorithmic}
\usepackage{graphicx}
\usepackage{algorithm,algorithmic}
\usepackage{hyperref}
\hypersetup{hidelinks}
\usepackage{textcomp}
\usepackage{booktabs}
\usepackage{siunitx}
\usepackage{multirow}
\usepackage{makecell}
\usepackage[english]{babel}
\usepackage[autostyle, english=american]{csquotes}
\MakeOuterQuote{"}
\usepackage[caption=true, font=footnotesize]{subfig}

\def\BibTeX{{\rm B\kern-.05em{\sc i\kern-.025em b}\kern-.08em
    T\kern-.1667em\lower.7ex\hbox{E}\kern-.125emX}}
\usepackage{bm}

\DeclarePairedDelimiter\autobracket{(}{)}
\newcommand{\br}[1]{\autobracket*{#1}}
\DeclarePairedDelimiter\sparen{[}{]}
\newcommand{\sbr}[1]{\sparen*{#1}}
\newcommand{\E}{\mathbb{E}}
\newcommand{\bx}{\bm{x}}

\newcommand{\KL}{\text{KL}}
\newcommand{\D}{\mathcal{D}}
\newcommand{\bz}{{\bm{\zeta}}}

\renewcommand{\Pr}{\text{Pr}}

\begin{document}

\title{Efficient Training of Probabilistic Neural Networks for Survival Analysis}

\author{Christian Marius Lillelund$^{1}$, Martin Magris$^{2}$ and Christian Fischer Pedersen$^{1}$%
\thanks{This work was supported by the PRECISE project (\url{http://www.aal-europe.eu/projects/precise/}) under Grant Agreement No. AAL-2021-8-90-CP by the European AAL Association.}
\thanks{$^{1}$The authors are with the Department of Electrical and Computer Engineering, Aarhus University, Finlandsgade 22, 8200 Aarhus N, Denmark. Emails: {\tt\small \{cl,cfp\}@ece.au.dk}. $^{2}$The author is with the Dipartimento di Scienze Economiche, Aziendali, Matematiche e Statistiche, Università degli Studi di Trieste, Via A. Valerio 4/1, 34127 Trieste, Italy. Email: martin.magris@deams.units.it}%
}
\maketitle
\begin{abstract}
Variational Inference (VI) is a commonly used technique for approximate Bayesian inference and uncertainty estimation in deep learning models, yet it comes at a computational cost, as it doubles the number of trainable parameters to represent uncertainty. This rapidly becomes challenging in high-dimensional settings and motivates the use of alternative techniques for inference, such as Monte Carlo Dropout (MCD) or Spectral-normalized Neural Gaussian Process (SNGP). However, such methods have seen little adoption in survival analysis, and VI remains the prevalent approach for training probabilistic neural networks. In this paper, we investigate how to train deep probabilistic survival models in large datasets without introducing additional overhead in model complexity. To achieve this, we adopt three probabilistic approaches, namely VI, MCD, and SNGP, and evaluate them in terms of their prediction performance, calibration performance, and model complexity. In the context of probabilistic survival analysis, we investigate whether non-VI techniques can offer comparable or possibly improved prediction performance and uncertainty calibration compared to VI. In the MIMIC-IV dataset, we find that MCD aligns with VI in terms of the concordance index (0.748 vs. 0.743) and mean absolute error (254.9 vs. 254.7) using hinge loss, while providing C-calibrated uncertainty estimates. Moreover, our SNGP implementation provides D-calibrated survival functions in all datasets compared to VI (4/4 vs. 2/4, respectively). Our work encourages the use of techniques alternative to VI for survival analysis in high-dimensional datasets, where computational efficiency and overhead are of concern.
\end{abstract}


\section{Introduction}

Uncertainty estimation is important in risk-sensitive applications, such as healthcare, where confidence in a prognostic outcome can have life-changing implications. Machine learning algorithms often make predictions that are subject to noise or inference errors \cite{gal_dropout_2016}, which motivates the use of models that can provide uncertainty estimates in their predictions and boost the point prediction delivered by traditional approaches. In this context, we refer to \textit{aleatoric} uncertainty, due to the inherent noise in the data, e.g., sampling noise, occlusions, or lack of quality features, and \textit{epistemic} uncertainty, due to lack of knowledge in the model, inversely proportional to the amount of training data \cite{magris_bayesian_2023}.

Survival analysis plays an important role in disease understanding and prognosis, and medical researchers use survival models to assess the importance of prognostic variables in outcomes such as death or cancer remission. Neural networks have been widely adopted for this task, e.g., DeepSurv \cite{katzman_deepsurv_2018}, Deep Survival Machines (DSM) \cite{nagpal_deep_2021} and Deep Cox Mixtures \cite{nagpal_deep_cox_2021}, but these methods cannot provide uncertainty estimates in their predictions. Bayesian Neural Networks (BNNs), i.e., neural networks estimated with Bayesian inference, have thus been adopted in survival analysis, using pseudo-probabilities \cite{feng_bdnnsurv_2021}, the CoxPH model \cite{qi_using_2023}, or implemented with the Multi-Task Logistic Regression (MTLR) framework \cite{loya2020uncertainty, qi_using_2023}, however, these approaches use the traditional Variational Inference (VI) technique to estimate the model parameters \cite{graves_practical_2011}, which, among the computational drawbacks, requires doubling the number of model parameters \cite{gal_dropout_2016}. This might not be feasible in high-dimensional settings, constituting a relevant limitation of VI. The use of alternative methods to train BNNs for survival analysis is limited, as the predominant approaches are based on VI \cite{feng_bdnnsurv_2021, loya2020uncertainty, qi_using_2023}, and no prior work in probabilistic survival analysis has compared VI with other approximation techniques in terms of prediction and calibration performance.

In this work, we adopt three competing approaches for training deep probabilistic survival models: VI, Monte Carlo Dropout (MCD), and Spectral-normalized Neural Gaussian Process (SNGP). We evaluate all three approaches in terms of prediction and calibration performance and compare them to established literature benchmarks. The goal is to determine whether two lightweight training techniques (MCD and SNGP) can provide the same degree of performance and uncertainty calibration as VI, but without the computational overhead. Our contributions are several\footnote{The code for replicating the results is available at: \\ \url{https://github.com/thecml/baysurv}.}: 


\begin{itemize}
    \item We adopt three competing approaches for training deep probabilistic survival models: VI, MCD, and SNGP.
    \item We show that MCD and SNGP are suitable alternatives to VI, as they both rival the prediction and calibration performance of VI, but without needing to double the number of model parameters.
    \item We evaluate the approaches in four real-world datasets.
\end{itemize}

\section{Related work}
\label{sec:related_work}

Probabilistic survival analysis using BNNs has recently acquired interest \cite{feng_bdnnsurv_2021, loya2020uncertainty, qi_using_2023, lillelund_uncertainty_2023}, as Bayesian models can express uncertainty about their predictions and provide implicit regularization via the prior \cite{magris_bayesian_2023}. In contrast to maximum likelihood estimation (MLE), a Bayesian model trained by maximum a posteriori averages predictions over a posterior distribution of the set of parameters: where there is no data, the confidence intervals will diverge, indicating that there are many possible extrapolations \cite{magris_bayesian_2023}.

Loya et al. \cite{loya2020uncertainty} proposed a Bayesian extension of the discrete MTLR framework that can capture patient-specific survival uncertainties, as a remedy for the assumption of temporal consistency of relative risk between two patients \cite{NIPS2011_1019c809}. MTLR uses multitask regression and joint likelihood minimization to model the log-risk at a given time interval as a linear combination of the covariates. The authors adopted this structure using a BNN with an element-wise structure (the first and second layers must have the same number of neurons) with feature selection in mind. Using VI to learn the posterior distribution and drawing samples from the predictive distribution, their model improves over traditional CoxPH and MTLR in predictive accuracy while providing confidence bounds on the predicted survival probability.

Qi et al. adopted the CoxPH and MTLR framework to formulate a BNN for survival analysis using VI \cite{qi_using_2023}. The authors investigated the selection of an appropriate prior distribution on the model weights and how this can enforce sparsity for later feature selection. They also proposed a novel framework for estimating individual survival distributions with credible intervals (CrIs) around the survival function, thus communicating the underlying uncertainty on the model weights. Their work is applicable as a decision support system, as predicted survival functions are specific to individual patients.

Lillelund et al. showed that MCD can provide satisfactory predictive results in survival models while being significantly faster to train than VI \cite{lillelund_uncertainty_2023}. Using the CoxPH framework, they quantified epistemic and aleatoric uncertainty by replacing the network's weights with a distribution and sampling the network’s outputs from a Gaussian. The results in Lillelund et al. \cite{lillelund_uncertainty_2023} demonstrate that aleatoric and epistemic uncertainty can be included in a deep survival model at some computational cost, but the authors do not report the calibration performance of the models or provide any measurement of the quality of the uncertainty estimates. In addition, the evaluation is performed only on a relatively small dataset (8873 samples).

Except for \cite{lillelund_uncertainty_2023}, all previous work uses VI \cite{graves_practical_2011} to infer the posterior distribution over the network parameters. This method approximates the true posterior distribution with a surrogate of a tractable parametric form. However, VI requires doubling (at least) the number of model parameters and comes with a high computational cost \cite{gal_dropout_2016}, e.g., because of the extensive use of simulation and gradient computation, which is restrictive and inefficient in high-dimensional settings. This motivates adopting non-VI techniques, but using such techniques to train deep survival models is underexplored.

\section{Methods}
\label{sec:methods}

This section introduces the reader to the main concepts in survival analysis, outlines the popular CoxPH model fundamental to our work, and describes three modern and commonly used approaches for training probabilistic models. Subsection \ref{subsec:proposed_architectures} discusses our proposed network architectures.

\subsection{Survival analysis}

Survival analysis models the time until an event occurs. This event can be observed or not, e.g., a censored observation due to the termination of the study. Let $\mathcal{D} = \left\lbrace \br{y_i, \delta_i, \bx_i} \right\rbrace_{i=1}^N$ be the data for the $i$th record (individual), where $y_i$ denotes the observed time, $\delta_{i}$ is a binary event indicator (the event occurred/did not occur) and $\bm{x}_{i} \in \mathbb{R}^{d}$ is a $d$-dimensional vector of features (covariates). Moreover, let $c_i$ denote the censoring time and $e_{i}$ denote the event time for the $i$th record, thus $y_i = e_i$ if $\delta_i = 1$ or $y_i = c_i$ if $\delta_i = 0$. Survival analysis then models the survival probability $S\br{t} = \Pr\br{T>t} = 1-\Pr\br{t\leq T}$, i.e., the probability that some event occurs at time $T$ later than $t$. Let $t_{k}$ denote some measurement time on some discretized event horizon of $K$ bins, $[t_0,\; t_1,\; t_2,\; \ldots,\; t_K]$, then at $t_{0}$ the survival function $S(t_{0})$ is one and monotonically decreases for $t_{k}>0$, so that the probability of surviving indefinitely is zero. In the absence of censored observations, $S(t_{k})$ corresponds to the fraction of survivors at time $t_{k}$. In survival analysis, the hazard function is commonly used to estimate $S\br{t}$:

\begin{equation}\label{eq:ht}
h\br{t} = \lim_{\Delta t \rightarrow 0} \Pr\br{t<T\leq t+ \Delta t \vert T>t}/\Delta t\text{,}
\end{equation}

\noindent which provides the instantaneous failure {\it rate} at given time instance $t$, conditional on surviving $t$ \cite[Ch. 11]{gareth_introduction_2021}. The hazard function and the survival function are connected through $h\br{t} = f\br{t}/S\br{t}$, with $f\br{t}$ being the probability density of $T$, i.e. $f\br{t} := \lim_{\Delta t \rightarrow 0} \Pr\br{t<T\leq t+\Delta t}/\Delta t $, the instantaneous (unconditional) rate of failure at $t$. The functions $S\br{t}$, $h\br{t}$, $f\br{t}$, thus summarize different probabilistic aspects of the random variable $T$, and are related to each other.

\subsection{The Cox Proportional Hazard model}

The Cox Proportional Hazards model (CoxPH) is a standard approach for survival analysis \cite{cox_regression_1972}. By adopting a multiplicative form for the contribution of several covariates to each individual's survival time, the CoxPH model is a powerful yet simple tool for assessing the simultaneous effect different covariates have on survival times. The conditional individual hazard function assumed by the CoxPH model is $h\br{t\vert \bx_i} = h_0\br{t} \exp \br{f\br{\bm{\omega},\bx_i}}$, where $h_0\br{t}$ stands for the baseline hazard at time $t$, $f\br{\bm{\omega},\bx_i}$ is called the risk score and $f$ is a linear function in the parameters \cite{cox_regression_1972}. Even if $h_0\br{t}$ remains unknown, it is possible to estimate $\bm{\omega}$ independently in the exponential part, with estimated hazards that satisfy the non-negative constraint. Its standard MLE of $\hat{\bm{\omega}}$ optimizes the Cox partial log-likelihood as follows:

\begin{align}\label{eq:cox_ll}
\log p\br{\D\vert \bm{\omega}} &= \sum_{i:\delta_i=1}\log \frac{h\br{y_i \vert \bx_i}}{ \sum_{j:T_j\geq T_i} h\br{y_i\vert \bx_j}}\text{,} \nonumber\\
 &= \sum_{i:\delta_i=1}f\br{\bm{\omega},\bx_i} - \sum_{j:T_j\geq T_i}f\br{\bm{\omega},\bx_j}\text{.}
\end{align}

Among the available estimators for $h_0\br{t}$, leading to the estimation of the survival function as $\hat{S}\br{t} = \hat{S_0}\br{t}^{\exp\br{\bx_i\hat{\bm{\omega}}}}$, with $\hat{S}_0\br{t} = \exp(-\int_0^t\hat{h}_0\br{u} \text{d} u)$, the Breslow estimator is the standard \cite{breslow_analysis_1975}.

\subsection{Uncertainty in Bayesian deep learning}\label{sec:VI}

In this subsection, we introduce three major approaches for training probabilistic machine learning models.

\textbf{Variational Inference (VI):} Let $\D$ denote the data, $p\br{\D\vert \bm{\omega}}$  the likelihood, and let the prior distribution on the parameter be denoted by $\bm{\omega}$ as $p\br{\bm{\omega}}$. The object of interest in Bayesian inference is the posterior distribution $p\br{\bm{\omega}\vert \D} =  p\br{\D\vert \bm{\omega}}p\br{\bm{\omega}}/p\br{\D}$. 
To obtain the distribution of the response $\bm{y}^\star$ for a new data sample $\bm{x}^\star$, the so-called posterior distribution is obtained as $p\br{\bm{y}^\star\vert \bm{x}^\star} = \int_\Theta p\br{\bm{y}^\star\vert \bm{x}^\star, \bm{\omega}}p\br{\bm{\omega} \vert \D} \text{d}\bm{\omega}$.

In general, determining the posterior is not an easy task, as the model evidence $p\br{\D}$ corresponds to an intractable integral. When the dimensionality of the problem is high, the use of sampling methods is prohibitive, and VI constitutes an attractive alternative. In VI, the actual unknown posterior distribution is approximated with a variational distribution $q\br{\bm{\omega}}$ of a tractable parametric form, whose (variational) parameter is denoted by $\bz$.
Typically, this corresponds to a multivariate Gaussian with mean $\bm{\mu}$ and covariance matrix $\bm{\Sigma}$, thus $\bz =[\bm{\mu};\text{vec}\,\bm{\Sigma}]$. VI achieves the best variational approximation to $p\br{\bm{\omega}\vert \D}$ by minimizing the KL divergence from $q\br{\bm{\omega}}$ to $p\br{\bm{\omega}\vert \D}$, 
i.e., determines the variational parameter $\bz^\star$ by minimizing:
\begin{align}\label{eq:kl}
\KL(q \br{\bm{\omega}} &\vert \vert p\br{\bm{\omega} \vert \D}) :=\nonumber\\
&-\int q\br{\bm{\omega}} \log p\br{\D \vert \bm{\omega}}\text{d}\bm{\omega} + \KL\br{q\br{\bm{\omega}} \vert \vert p\br{\bm{\omega}}}\text{,}
\end{align}
for which several algorithms have been developed in the last decade \cite{magris_bayesian_2023}.

\textbf{Monte-Carlo Dropout (MCD):} \cite{gal_dropout_2016} and \cite{gal2016uncertainty} provide a formal connection between Monte Carlo Dropout (MCD) and BNNs. Here, we discuss the major breakthrough for the relevant non-categorical case for the $\bm{y}$s. Consider a deep Gaussian Process (GP) with a covariance function of the form:
\begin{align}
\bm{K}\br{\bm{x},\bm{y}}=\int p\br{\bm{w}} p\br{b} \sigma\br{\bm{w}' \bm{x}+b}\sigma\br{\bm{w}' \bm{y}+b}\text{d} \bm{w} \text{d}b\text{,}
\end{align}
\noindent where $\bm{W}_i$ is the weight matrix of dimension $K_i \times K_{i-1}$ for each layer $i = 1,\dots,L$ and $\bm{b}_i$ is the corresponding bias vector. We use $\bm{\omega} = \{\bm{W}_i\}_{i=1}^L$, where each row of $\bm{W}_i$ is distributed according to a standard multivariate normal distribution $p\br{\bm{w}} = \mathcal{N}\br{\bm{w};\bm{0},l^{-2}\bm{I}}$. Let $\sigma\br{\cdot}$ be a general non-linear activation function and $p\br{b}$ some distribution on the elements of $\bm{b}_i$. Given some precision parameter $\tau>0$, the predictive distribution of the deep GP is
\begin{align}
p\br{\bm{y} \vert \bm{x}} = \int p\br{\bm{y}\vert \bm{x}, \bm{\omega}}p\br{\bm{\omega}\vert \D} \text{d} \bm{\omega}\text{,}
\end{align}
with
\begin{align}
p\br{\bm{y}\vert \bm{x}, \bm{\omega}} = \mathcal{N}\br{\bm{y};\bm{\hat{y}}\br{\bm{x},\bm{\omega}},\tau^{-1}\bm{I}}\text{.}
\end{align}
For the intractable posterior $p\br{\bm{\omega}\vert \D}$, we adopt a variational approximation $q\br{\bm{\omega}}$ defined as:
\begin{align}
\bm{W}_i &= \bm{M} \cdot \text{diag}\br{\sbr{z_{i,j}}_{j=1}^{K_i}}\text{,}\\
z_{i,j} &\sim \text{Bernoulli}\br{p_i},\,i=1,\dots,L\text{,}\,\,j=1,\dots,K_{i-1}\text{,}
\end{align}
\noindent where $\bm{M}_i$ are the variational parameters and $p_i$ some given probabilities. \cite{gal_dropout_2016} shows that minimizing the KL divergence between the variational posterior $q\br{\bm{\omega}}$ and the actual posterior of the deep GP is equivalent to optimizing the following objective with dropout:
\begin{align}
\mathcal{L}_\text{MCD} = \frac{1}{N}\sum_{i=}^N E\br{\bm{y}_i,\bm{\hat{y}}_i} +\lambda\sum_{i=1}^N \br{\vert \vert\bm{W}_i \vert\vert_2^2 + \vert\vert \bm{b}_i \vert\vert_2^2 }\text{.}
\end{align}
$L$ is a standard non-probabilistic optimization objective, using the $L_2$ regularization and tuning parameter $\lambda$ and the Euclidean square loss $E\br{\cdot,\cdot}$. Dropout means that for every input data and every unit in each layer, a Bernoulli variable is sampled with probability $p_i$, and a unit is dropped if the corresponding binary variable is 0. The probability $1-p$ of dropping a unit is generally called the dropout rate. Therefore, a DNN, with dropout applied at every layer, is equivalent to a Monte Carlo approximation to the probabilistic deep GP; i.e., it minimizes the KL divergence between the variational distribution and the posterior of the deep GP modeling the distribution over the space of distributions that can have generated the data. It is straightforward to approximate the predictive distribution as follows:
\begin{align}
    p\br{\bm{y}^\star \vert \bm{x}^\star} &= p\br{\bm{y}^\star \vert \bm{x}^\star,\bm{\omega}} q\br{\bm{\omega}} \text{,}\\
    q\br{\bm{\omega}} &= \text{Bernoulli}\br{z_1} \ldots \text{Bernoulli}\br{z_L}\text{,}\\
    p\br{\bm{y}^\star \vert \bm{x}^\star,\bm{\omega}} &= \mathcal{N}\br{\bm{y}^\star;\bm{\hat{y}}^\star\br{\bm{x}^\star,z_1,\dots,z_L}, \tau^{-1}\bm{I} }\text{,}
\end{align}
and, e.g., its mean through MC sampling:
\begin{align}
\E_{p\br{\bm{y}^\star \vert \bm{x}^\star}} &\approx \frac{1}{S}\sum_{s=1}^S \hat{y}^\star\br{\bm{x}^\star,\hat{z}_{1,s},\dots,\hat{z}_{L,s}}\text{,}\\
\hat{z}_{i,s} &\sim  \text{Bernoulli}\br{p_i}\text{.}
\end{align}
The practical simplicity of MCD makes it an attractive and widely adopted baseline for comparison with alternative Bayesian approaches. However, within neural network applications in survival analysis, the use of MCD for uncertainty quantification is very limited \cite{lillelund_uncertainty_2023}.

\textbf{Spectral-normalized Neural Gaussian Process (SNGP):} SNGP \cite{liu2020simple} is an approach for improving the quality of uncertainty modeling in a deep neural network whilst preserving its overall predictive accuracy. Over a deep residual network, SNGP applies spectral normalization to the hidden layers and replaces the dense output layer with a GP layer. Compared to MCD, SNGP works for a wide range of state-of-the-art residual-based architectures. Furthermore, it does not rely on ensemble averaging and, therefore, has a similar level of latency as a deterministic network (besides requiring a covariance reinitialization of low complexity at each iteration), scaling easily to large dimensions and lengthy datasets \cite{liu2020simple}.

\subsection{Proposed architectures}
\label{subsec:proposed_architectures}

\begin{figure*}[!htbp]
\centering
\subfloat[VI.\label{fig:BNN}]{%
\includegraphics[trim={2cm 1.0cm 0.8cm 1cm},clip,width=0.5\textwidth]{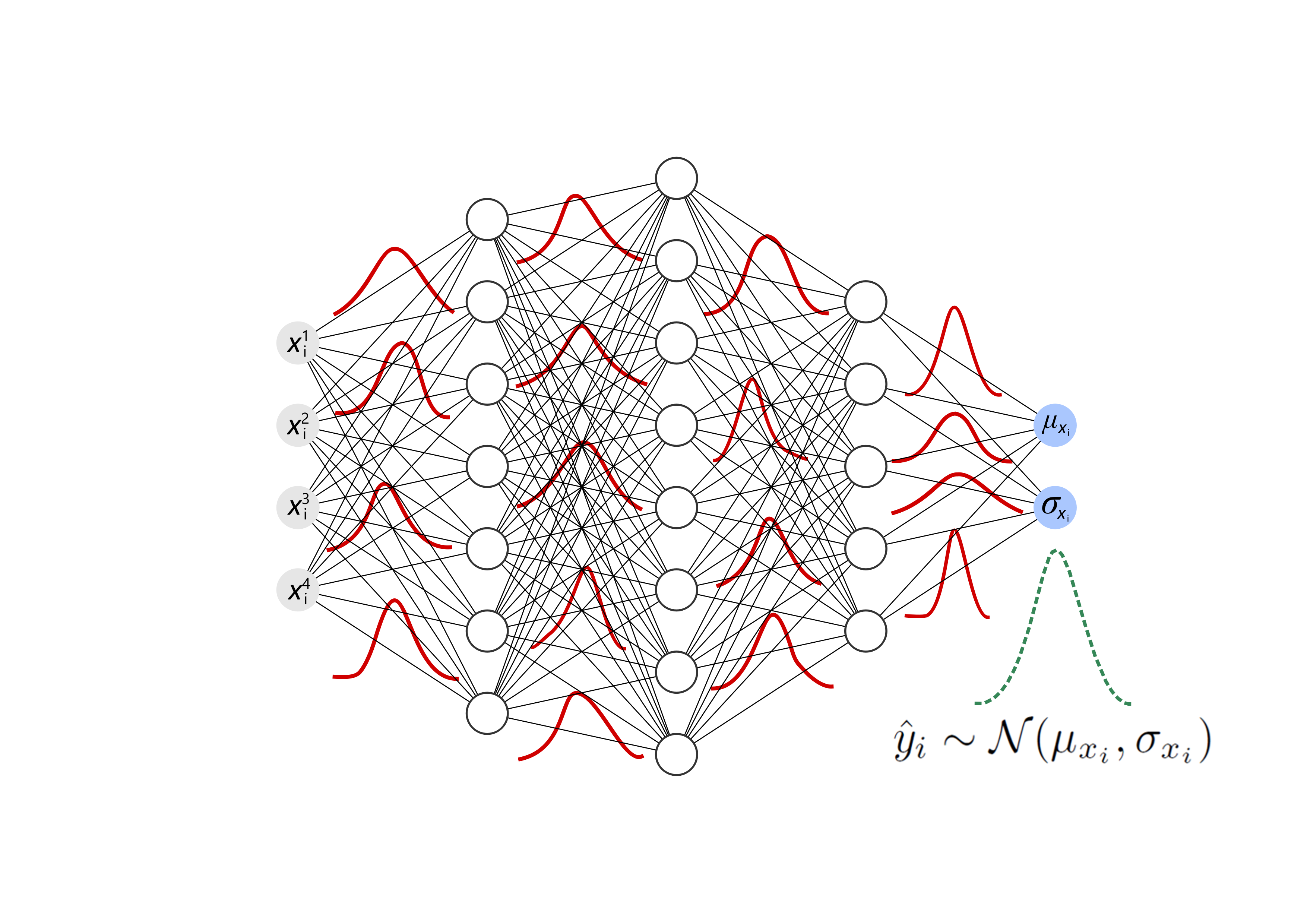}}
\subfloat[MCD.\label{fig:MCD}]{%
\includegraphics[trim={2cm 1.0cm 0.8cm 1cm},clip,width=0.5\textwidth]{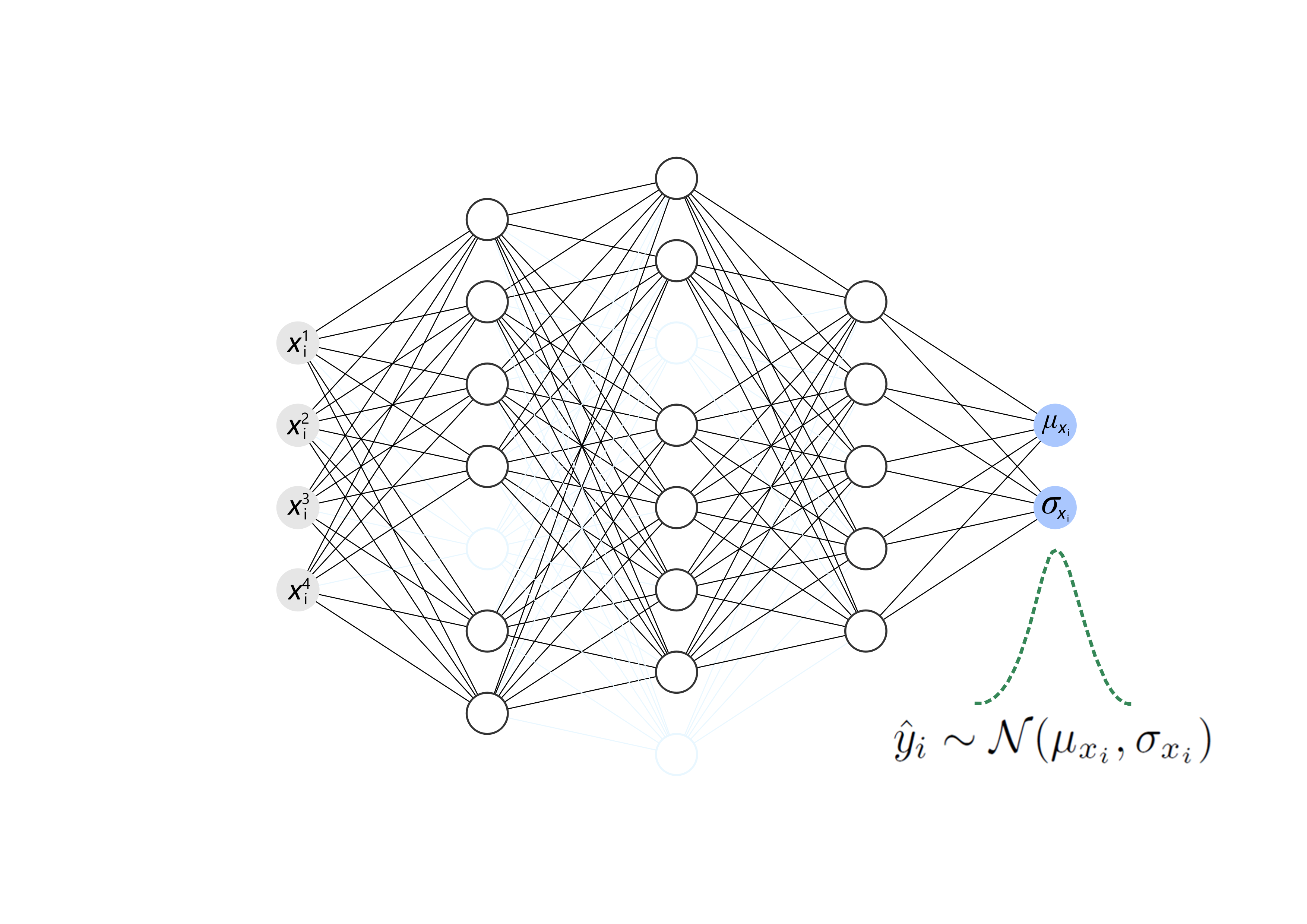}}
\\
\subfloat[SNGP.\label{fig:SNGP}]{%
\includegraphics[trim={0cm 2.5cm 0cm 2.5cm},clip,width=0.5\textwidth]{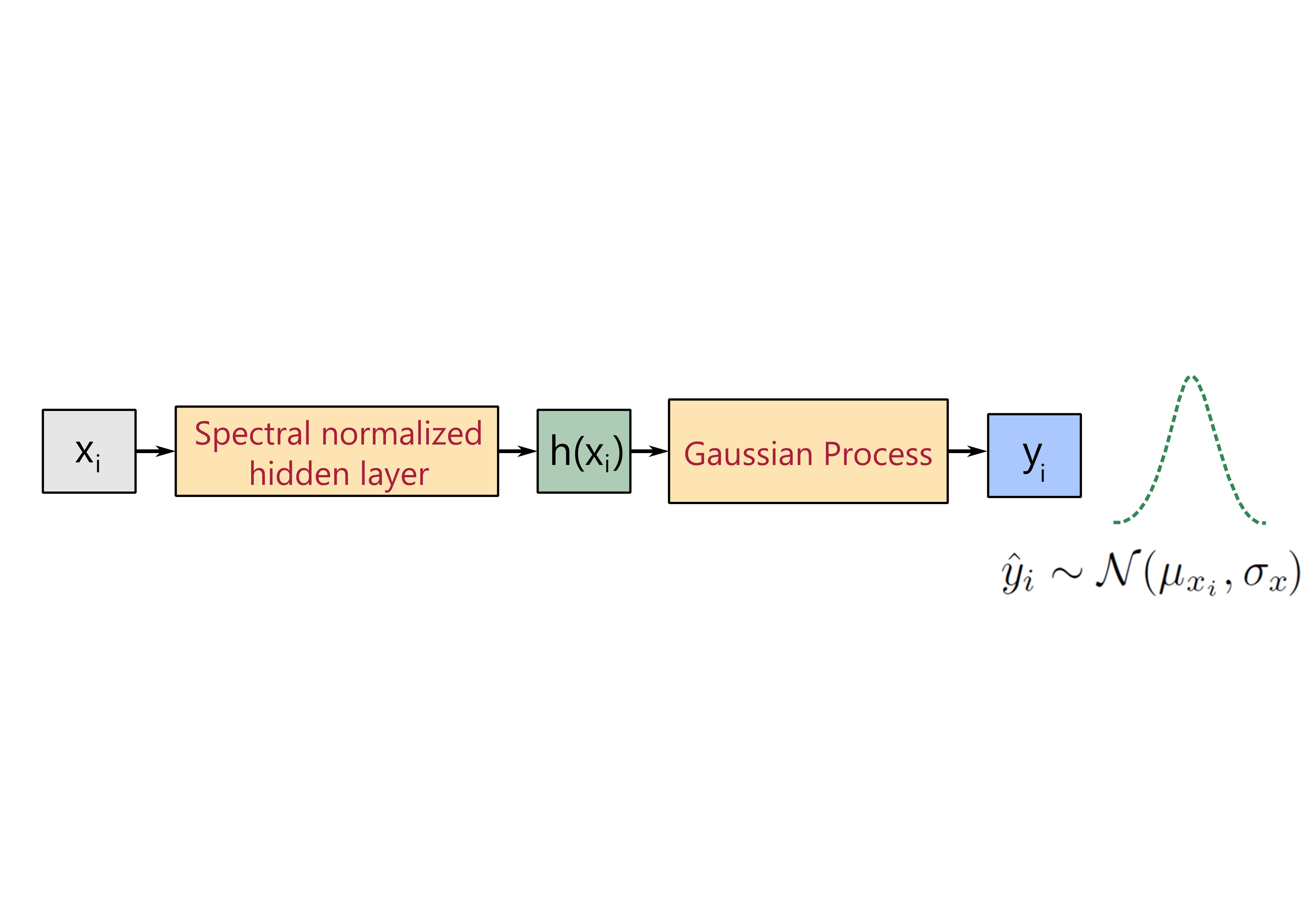}}
\caption{The proposed architectures for computing $\hat{y}_{i}$, i.e., the predicted survival time for individual $i$. (a): BNN with three hidden layers trained with VI, where $\bm{x}_i \in \mathbb{R}^4$ is a four-dimensional covariate vector. (b): MLP with three hidden layers trained with MCD. Transparent objects emphasize the random activation of the nodes under MCD. (c): MLP using the SNGP technique, with spectral normalized hidden layers and the output computed from a GP layer.}
\label{fig:NNArchitectures}
\end{figure*}

We propose three separate neural network architectures for probabilistic survival analysis. For the purpose of comparability in the results, consistency in the hyperparameters, and evaluation fairness, we propose a Multi-Layer Perceptron (MLP) as the backbone network architecture. An MLP is a fully connected feedforward network; the number of hidden layers and neurons are hyperparameters tuned based on the data, see, e.g., \cite{magris_bayesian_2023}. The adoption of an MLP is aligned with the literature, as capable of providing solid prediction performance \cite{katzman_deepsurv_2018, nagpal_deep_2021}. The proposed architectures are depicted in Fig. \ref{fig:NNArchitectures}. There are three probabilistic training approaches:

\begin{itemize}
\item[i.] training with VI by optimizing Eq. \eqref{eq:kl}, where the weights and biases are treated as random variables. The output nodes provide the mean and standard deviation of the output $\hat{y}_i$ as a Gaussian draw, i.e., $\hat{y}_i \sim \mathcal{N}\br{\mu_{\bm{x}_i},\sigma_{\bm{x}_i}}$ for a $d$-dimensional sample $\bm{x}_i$. The Bayesian framework captures the epistemic uncertainty in the model, and sampling the output from a Gaussian captures the aleatoric uncertainty. See Fig. \ref{fig:BNN}.
\item[ii.] training with MCD, capturing both epistemic and aleatoric uncertainty. 
The network architecture is equivalent to the deterministic MLP, but now MC dropout is applied at each feedforward pass. The two output nodes correspond to the mean and variance of the predictive Gaussian, from which the $y$s are drawn conditionally on $\bm{x}_i$ and the random network configuration at each epoch due to the random deactivation of some of the nodes. 
We use dropout rates $p=\{0.1, 0.2, 0.5\}$. See Fig. \ref{fig:MCD}.
\item[iii.] training with SNGP, which can estimate the uncertainty per sample without requiring random sampling. Given a batch input $\{\bm{x}_1,\bm{x}_{2}, ..., \bm{x}_{N}\}$, the GP layer returns predictions $\{y_1, y_{2}, ..., y_{N}\}$ along with the predictive covariance matrix of the predictions. The model is optimized according to the MLE objective in Eq. \eqref{eq:cox_ll}. See Fig. \ref{fig:SNGP}.
\end{itemize}

Table \ref{tab:model_architectures} highlights the advantages and disadvantages of adopting VI, MCD and SNGP.

\begin{table*}[htbp]
\centering
\caption{Advantages and disadvantages of the three proposed architectures.}
\label{tab:model_architectures}
\begin{tabular}{@{}lcc@{}}
\toprule
 & \textit{Advantages:} & \textit{Disadvantages:} \\ \cmidrule(lr){2-2} \cmidrule(lr){3-3}
VI \cite{graves_practical_2011} & {\makecell{+ Mathematically-robust \& tractable\\+ Provides a posterior distribution}} & {\makecell{- Requires training additional model parameters\\-Heavily relies on MC sampling\\- Has long training times compared to MCD and SNGP \cite{gal_dropout_2016,liu2020simple}}} \\
\cmidrule(lr){2-3}
MCD \cite{gal_dropout_2016} & {\makecell{+ No parametric overhead compared to VI \\+ Has shorter training times than VI on average}} & {\makecell{- Does not provide an actual posterior distribution \\ - Only approximates the VI solution under certain assumptions \cite{gal_dropout_2016} \\ - Uncertainty estimation coupled to the choice of dropout rate \cite{NIPS2016_8d8818c8}}} \\
\cmidrule(lr){2-3}
SNGP \cite{liu2020simple} & {\makecell{+ Fast, deterministic approach \\ + No parametric overhead compared to VI \\+ Has shorter training times than VI on average}} & {\makecell{- Does not provide an actual posterior distribution \\ - Cannot compute individual CrIs}} \\ \bottomrule 
\end{tabular}
\end{table*}

\section{Experiments and results}
\label{sec:experiments}

\subsection{Data and setup}

We use four datasets to evaluate the proposed architectures. (1) The Molecular Taxonomy of Breast Cancer International Consortium (METABRIC) dataset contains gene and protein expression profiles used to identify breast cancer subgroups \cite{metabric_group_genomic_2012}. We obtain the dataset from \cite{katzman_deepsurv_2018}, consisting of 1904 observations, 9 covariates, and a censoring rate of 42\%. (2) The U.S. Surveillance, Epidemiology, and End Results (SEER) dataset contains survival times for cancer patients \cite{ries_cancer_2007}, specifically patients with infiltrating duct and lobular carcinoma breast cancer diagnosed between 2006 and 2010. The dataset has 4024 observations and a censoring rate of 85\%. (3) The Study to Understand Prognoses Preferences Outcomes and Risks of Treatment (SUPPORT) dataset is a large study that tracks the survival time of hospitalized adults \cite{knaus_support_1995}; the cleaned dataset is available from \cite{katzman_deepsurv_2018} and consists of 8873 observations, 14 covariates, and a censoring rate of 32\%. (4) MIMIC-IV is a large electronic health record dataset with 40,000 patients admitted to intensive care units at the Beth Israel Deaconess Medical Center in the United States between 2008 and 2019 \cite{johnson_mimic-iv_2023}. The dataset includes patient demographics, clinical measurements, medications, and diagnoses, recorded during a patient's stay at the hospital and their survival time. See \cite{qi_using_2023} for accessing the data, which has 38,520 observations, 91 covariates, and a censoring rate of 67\%. After imputing missing values by sample mean for real-valued covariates or mode for categorical covariates, we apply a $z$-score data normalization and use one-hot encoding for categorical covariates. We split the data into train, validation and test sets by 70\%, 10\%, and 20\% using a stratified procedure, as proposed by \cite{qi_using_2023}. The stratification ensures that the event times and censoring rates are consistent across the three sets.

For the purpose of comparability in the results, we implement the traditional continuous CoxPH model using a linear estimator \cite{cox_regression_1972}, weight regularization \cite{simon_regularization_2011}, gradient boosting \cite{friedman_greedy_2001}, Random Survival Forests \cite{ishwaran_random_2008}, two non-Bayesian neural networks \cite{nagpal_deep_2021,nagpal_deep_cox_2021}, and two recent BNNs based on the continuous CoxPH model and the discrete MTLR framework, respectively \cite{qi_using_2023}. Our baseline MLP is equivalent to the DeepSurv model \cite{katzman_deepsurv_2018}. For all datasets and models, we use Bayesian optimization \cite{snoek_practical_2012} to tune the hyperparameters over ten iterations using the validation set. These include the number of iterations, the batch size, and the network architecture, when applicable. We used the hyperparameters leading to the highest average concordance index to configure the final models. The CoxPH-based models utilize the entire dataset during training, given the nature of the partial likelihood, whereas the neural networks load stochastic subsets of the training set as mini-batches. Additionally, we use the early stopping technique, where the training process halts if the likelihood loss of the model stops improving on the validation dataset, for all neural network-based models. See App. \ref{app:model_hyperparameters} for more details on hyperparameter optimization.

We adopt several evaluation metrics to assess the prediction and calibration performance of our baseline MLP model, its probabilistic variants, and literature benchmarks. To evaluate prediction performance, we report Antolini's Concordance Index (CI\textsubscript{td}) \cite{antolini_time-dependent_2005}, the Mean Absolute Error using hinge loss (MAE\textsubscript{H}) \cite{qi_survivaleval_2024}, the MAE using pseudo-observations (MAE\textsubscript{PO}) \cite{qi_survivaleval_2024}, and the Integrated Brier Score (IBS) \cite{graf_assessment_1999}. To evaluate calibration performance, we report the Integrated Calibration Index (ICI) \cite{austin_ici_2019}, Distribution Calibration (D-Cal) \cite{haider_effective_2020} and Coverage Calibration (C-Cal) \cite{qi_using_2023}. See App. \ref{app:performance_metrics} for more details on the performance metrics.

\subsection{Prediction results} \label{sec:prediction_results}

\begin{table*}[!htbp]
\centering
\caption{Prediction performance on the METABRIC, SEER, SUPPORT, and MIMIC-IV test sets. In many cases, the proposed models give notable improvements in mean absolute error over continuous state-of-the-art baselines, with similar scores between VI and MCD. Bold indicates the top 3 models including ties. $N$: sample size, $C$: pct. of censored data, $d$: number of covariates.}
\label{tab:prediction_results}
\begin{minipage}{0.48\textwidth}
\centering
\subfloat[METABRIC ($N=1902$, $C=42\%$, $d=9$).\label{subtab:prediction_metabric}]{%
\begin{tabular}{lllll} 
{Model} & {CI\textsubscript{td} $\uparrow$} & {MAE\textsubscript{H} $\downarrow$} & {MAE\textsubscript{PO} $\downarrow$} & {IBS $\downarrow$} \\ \midrule
CoxPH \cite{cox_regression_1972} & 0.618 & 68.923 & 96.797 & 0.173 \\
CoxNet \cite{simon_regularization_2011} & 0.611 & 68.288 & 97.631 & 0.173 \\
CoxBoost \cite{hothorn_survival_2005} & 0.638 & 69.976 & 102.745 & 0.187 \\
RSF \cite{ishwaran_random_2008} & 0.641 & 67.242 & 98.728 & 0.175 \\
DSM \cite{nagpal_deep_2021} & \textbf{0.670} & 70.361 & 100.83 & 0.170 \\
DCM \cite{nagpal_deep_cox_2021} & 0.637 & 74.187 & 104.532 & 0.179 \\
BNN-Cox (VI) \cite{qi_using_2023} & \textbf{0.672} & 65.777 & 94.184 & \textbf{0.166} \\
BNN-MTLR (VI) \cite{qi_using_2023} & \textbf{0.655} & \textbf{59.027} & \textbf{89.387} & \textbf{0.163} \\ \midrule
\textbf{\textit{This work:}}\\
Baseline (MLP) & 0.604 & 68.668 & 95.540 & 0.168 \\
+ VI & 0.634 & \textbf{63.776} & \textbf{91.843} & 0.168 \\
+ MCD ($p=0.1$) & 0.621 & 67.235 & 93.981 & 0.167 \\
+ MCD ($p=0.2$) & 0.628 & 67.371 & 96.184 & \textbf{0.166} \\
+ MCD ($p=0.5$) & 0.632 & \textbf{65.651} & \textbf{92.353} & \textbf{0.164} \\
+ SNGP & 0.631 & 67.378 & 93.434 & \textbf{0.166} \\ \bottomrule
\end{tabular}}
\end{minipage}%
\begin{minipage}{0.48\textwidth}
\centering

\subfloat[SEER ($N=4024$, $C=85\%$, $d=28$).\label{subtab:prediction_seer}]{%
\begin{tabular}{lllll} 
{Model} & {CI\textsubscript{td} $\uparrow$} & {MAE\textsubscript{H} $\downarrow$} & {MAE\textsubscript{PO} $\downarrow$} & {IBS $\downarrow$} \\ \midrule
CoxPH \cite{cox_regression_1972} & \textbf{0.723} & 100.849 & 196.898 & \textbf{0.083} \\
CoxNet \cite{simon_regularization_2011} & \textbf{0.721} & 95.253 & 197.066 & 0.083 \\
CoxBoost \cite{hothorn_survival_2005} & 0.689 & 106.635 & 232.204 & 0.092 \\
RSF \cite{ishwaran_random_2008} & 0.682 & 102.062 & 206.41 & 0.086 \\
DSM \cite{nagpal_deep_2021} & 0.699 & 153.84 & 413.543 & 0.088 \\
DCM \cite{nagpal_deep_cox_2021} & 0.713 & 108.464 & 223.397 & \textbf{0.082} \\
BNN-Cox (VI) \cite{qi_using_2023} & \textbf{0.716} & 95.184 & \textbf{202.151} & \textbf{0.081} \\
BNN-MTLR (VI) \cite{qi_using_2023} & 0.702 & \textbf{48.166} & \textbf{201.856} & 0.085 \\ \midrule
\textbf{\textit{This work:}}\\
Baseline (MLP) & 0.703 & 101.972 & 204.408 & 0.084 \\
+ VI & 0.699 & \textbf{81.923} & \textbf{199.796} & 0.084 \\
+ MCD ($p=0.1$) & 0.702 & 91.202 & 203.457 & 0.084 \\
+ MCD ($p=0.2$) & 0.691 & 89.103 & 203.323 & 0.085 \\
+ MCD ($p=0.5$) & 0.692 & \textbf{82.001} & 207.956 & 0.087 \\
+ SNGP & 0.698 & 106.032 & 209.498 & 0.084 \\ \bottomrule
\end{tabular}
}
\end{minipage}

\vspace*{0.5cm}

\begin{minipage}{0.48\textwidth}
\centering

\subfloat[SUPPORT ($N=8873$, $C=32\%$, $d=14$).\label{subtab:prediction_support}]{%
\begin{tabular}{lllll} 
{Model} & {CI\textsubscript{td} $\uparrow$} & {MAE\textsubscript{H} $\downarrow$} & {MAE\textsubscript{PO} $\downarrow$} & {IBS $\downarrow$} \\ \midrule
CoxPH \cite{cox_regression_1972} & 0.568 & 409.947 & 754.978 & 0.201 \\
CoxNet \cite{simon_regularization_2011} & 0.568 & 409.915 & 754.960 & 0.201 \\
CoxBoost \cite{hothorn_survival_2005} & \textbf{0.612} & \textbf{366.928} & \textbf{721.443} & 0.190 \\
RSF \cite{ishwaran_random_2008} & 0.602 & 377.862 & 736.800 & 0.200 \\
DSM \cite{nagpal_deep_2021} & \textbf{0.626} & 384.349 & 723.002 & \textbf{0.185} \\
DCM \cite{nagpal_deep_cox_2021} & \textbf{0.629} & 434.376 & 759.509 & \textbf{0.185} \\
BNN-Cox (VI) \cite{qi_using_2023} & 0.610 & 416.650 & 738.350 & 0.189 \\
BNN-MTLR (VI) \cite{qi_using_2023} & 0.606 & \textbf{369.060} & \textbf{707.679} & 0.186 \\ \midrule
\textbf{\textit{This work:}}\\
Baseline (MLP) & 0.609 & 391.573 & 722.709 & 0.187 \\
+ VI & 0.531 & 414.133 & 768.599 & 0.208 \\
+ MCD ($p=0.1$) & 0.600 & 416.175 & 738.567 & 0.193 \\
+ MCD ($p=0.2$) & 0.567 & 400.064 & 751.343 & 0.202 \\
+ MCD ($p=0.5$) & 0.605 & \textbf{374.046} & \textbf{717.227} & 0.192 \\
+ SNGP & 0.610 & 397.877 & 726.725 & \textbf{0.184} \\ \bottomrule
\end{tabular}}
\end{minipage}%
\begin{minipage}{0.48\textwidth}
\centering
\subfloat[MIMIC-IV ($N=38520$, $C=67\%$, $d=91$).\label{subtab:prediction_mimic}]{%
\begin{tabular}{lllll} 
{Model} & {CI\textsubscript{td} $\uparrow$} & {MAE\textsubscript{H} $\downarrow$} & {MAE\textsubscript{PO} $\downarrow$} & {IBS $\downarrow$} \\ \midrule
CoxPH \cite{cox_regression_1972} & 0.726 & 257.364 & 286.516 & \textbf{0.174} \\
CoxNet \cite{simon_regularization_2011} & 0.727 & 255.802 & 283.951 & \textbf{0.174} \\
CoxBoost \cite{hothorn_survival_2005} & 0.656 & 263.836 & 297.763 & 0.194 \\
RSF \cite{ishwaran_random_2008} & 0.694 & 260.668 & 293.247 & 0.188 \\
DSM \cite{nagpal_deep_2021} & 0.727 & 259.110 & \textbf{280.361} & 0.180 \\
DCM \cite{nagpal_deep_cox_2021} & 0.742 & 257.801 & 286.580 & \textbf{0.174} \\
BNN-Cox (VI) \cite{qi_using_2023} & 0.743 & 256.007 & 288.814 & \textbf{0.169} \\
BNN-MTLR (VI) \cite{qi_using_2023} & 0.744 & \textbf{250.605} & \textbf{278.710} & \textbf{0.167} \\ \midrule
\textbf{\textit{This work:}}\\
Baseline (MLP) & 0.744 & 256.694 & 283.635 & 0.178 \\
+ VI & 0.743 & 254.735 & 282.345 & 0.180 \\
+ MCD ($p=0.1$) & \textbf{0.746} & 255.815 & 283.672 & 0.177 \\
+ MCD ($p=0.2$) & \textbf{0.748} & \textbf{254.560} & 283.156 & 0.175 \\
+ MCD ($p=0.5$) & \textbf{0.748} & \textbf{254.971} & \textbf{281.928} & 0.177 \\
+ SNGP & 0.736 & 257.195 & 282.082 & 0.182 \\ \bottomrule
\end{tabular}}
\end{minipage}
\end{table*}

Table \ref{tab:prediction_results} shows the prediction performance of our baseline MLP, its probabilistic variants, and the literature benchmarks on the test sets. CI\textsubscript{td} measures the rank correctness between pairwise observations ($i$, $j$), i.e., if individual $i$ experiences the event before $j$, their predicted survival time should be shorter. MAE is the mean absolute error between the predicted and actual survival times, using either hinge loss (MAE\textsubscript{H}) or pseudo-observations (MAE\textsubscript{PO}) to calculate the error for censored individuals. We use the median of the survival function as the predicted survival time. IBS is the mean squared error between the binary event outcome $y \in \{0, 1\}$ and the predicted probability of the event over the entire event horizon.

For VI and MCD, the reported metrics are constructed based on predictive means by sampling 100 times from the posterior predictive distribution. We train the networks for 100 epochs using early stopping with five epochs of patience. Concerning the baseline MLP model and the probabilistic variants, in the smallest METABRIC dataset (see Table \ref{subtab:prediction_metabric}), all probabilistic models report a higher CI\textsubscript{td} than the baseline,  and MCD in particular shows significantly lower MAE\textsubscript{H} and MAE\textsubscript{PO}. Thus, using VI or MCD improves the prediction performance over the MLP in this dataset. In the mid-sized SEER dataset (see Table \ref{subtab:prediction_seer}), all probabilistic variants give approximately the same CI\textsubscript{td}, but have significantly lower MAE\textsubscript{H} than the baseline. In the larger SUPPORT dataset (see Table \ref{subtab:prediction_support}), only MCD outperforms both the baseline and its rival probabilistic models. In the large MIMIC-IV dataset (see Table \ref{subtab:prediction_mimic}), the effect of the different estimation approaches is reduced, and all measures are generally aligned, although VI and MCD still have an edge when it comes to MAE\textsubscript{H}. We see that for all datasets in this experiment, using MCD with $p=0.5$ leads to better convergence and performance than the baseline.

With respect to the existing models, MCD with $p=0.5$ outperforms all continuous models in terms of MAE\textsubscript{H} in 3/4 datasets and in terms of MAE\textsubscript{PO} in 2/4 datasets. In the large MIMIC-IV dataset, it outperforms all competing solutions in terms of CI\textsubscript{td}. Although BNN-MTLR does demonstrate even better MAE throughout, it is a discrete model with two important caveats: (1) The MTLR framework requires training $K$ logistic regression models, where $K$ is the number of discrete time bins, leading to a space complexity of $O(Kd)$ instead of $O(d)$ for the continuous models \cite{NIPS2011_1019c809} (see Table \ref{tab:model_complexity}). (2) The model is trained using VI, which itself doubles the number of parameters compared to the MLP baseline and MCD. Overall, we see an advantage in combining the MLP architecture with Bayesian approximation techniques to rank individuals at risk and predict the time to event. These benefits are most apparent in small-sized datasets, as the Bayesian algorithms naturally incorporate a form of regularization (the prior) and hence are less prone to overfitting. As the size of the dataset increases, the effect of the prior diminishes, and the Bayesian solution approaches the MLE. Most importantly, the proposed architectures most often do not deteriorate prediction performance in our experiments; rather, they often improve it.

\subsection{Calibration results}
\label{sec:calibration_results}

\begin{table*}[!htbp]
\centering
\caption{Calibration performance on the METABRIC, SEER, SUPPORT, and MIMIC-IV test sets. MCD and SNGP have better calibration performances (D-Cal and C-Cal) than VI in small datasets, but similar performances in larger ones. Bold indicates the top 3 models including ties. $N$: sample size, $C$: pct. of censored data, $d$: number of covariates.}
\label{tab:calibration_results}
\begin{minipage}{0.48\textwidth}
\centering
\subfloat[METABRIC ($N=1902$, $C=42\%$, $d=9$).\label{subtab:calibration_metabric}]{%
\begin{tabular}{lccc} 
{Model} & {ICI $\downarrow$} & {D-Cal $\uparrow$} & {C-Cal $\uparrow$} \\ \midrule
CoxPH \cite{cox_regression_1972} & 0.022 & \textbf{0.959}  & - \\
CoxNet \cite{simon_regularization_2011} & \textbf{0.011} & \textbf{0.977}  & - \\
CoxBoost \cite{hothorn_survival_2005} & 0.147 & 0.822  & - \\
RSF \cite{ishwaran_random_2008} & 0.047 & 0.530  & - \\
DSM \cite{nagpal_deep_2021} & 0.025 & 0.017  & - \\
DCM \cite{nagpal_deep_cox_2021} & 0.066 & 0.160  & - \\
BNN-Cox (VI) \cite{qi_using_2023} & \textbf{0.017} & 0.101  & 0.812  \\
BNN-MTLR (VI) \cite{qi_using_2023} & 0.044 & 0.438  & 0.763  \\ \midrule
\textbf{\textit{This work:}}\\
Baseline (MLP) & 0.040 & 0.853  & - \\
+ VI & 0.062 & 0.059  & 0.796  \\
+ MCD ($p=0.1$) & \textbf{0.017} & 0.735  & \textbf{0.849}  \\
+ MCD ($p=0.2$) & 0.020 & 0.689  & \textbf{0.857}  \\
+ MCD ($p=0.5$) & \textbf{0.017} & 0.760  & \textbf{0.865}  \\
+ SNGP & \textbf{0.017} & \textbf{0.964}  & - \\ \bottomrule
\end{tabular}}
\end{minipage}%
\begin{minipage}{0.48\textwidth}
\centering
\subfloat[SEER ($N=4024$, $C=85\%$, $d=28$).\label{subtab:calibration_seer}]{%
\begin{tabular}{lccc} 
{Model} & {ICI $\downarrow$} & {D-Cal $\uparrow$} & {C-Cal $\uparrow$} \\ \midrule
CoxPH \cite{cox_regression_1972} & 0.011 & 0.999  & - \\
CoxNet \cite{simon_regularization_2011} & 0.013 & \textbf{1.000}  & - \\
CoxBoost \cite{hothorn_survival_2005} & 0.051 & \textbf{1.000}  & - \\
RSF \cite{ishwaran_random_2008} & 0.021 & \textbf{1.000}  & - \\
DSM \cite{nagpal_deep_2021} & \textbf{0.009} & 0.931  & - \\
DCM \cite{nagpal_deep_cox_2021} & 0.016 & \textbf{1.000}  & - \\
BNN-Cox (VI) \cite{qi_using_2023} & \textbf{0.009} & \textbf{1.000}  & \textbf{0.489}  \\
BNN-MTLR (VI) \cite{qi_using_2023} & 0.056 & 0.016  & 0.486  \\ \midrule
\textbf{\textit{This work:}}\\
Baseline (MLP) & \textbf{0.009} & \textbf{1.000}  & - \\
+ VI & 0.027 & 0.678  & \textbf{0.487}  \\
+ MCD ($p=0.1$) & 0.013 & 0.989  & \textbf{0.487}  \\
+ MCD ($p=0.2$) & 0.014 & 0.983  & \textbf{0.487}  \\
+ MCD ($p=0.5$) & 0.037 & 0.841  & 0.486  \\
+ SNGP & 0.011 & \textbf{1.000}  & - \\ \bottomrule
\end{tabular}}
\end{minipage}

\vspace*{0.5cm}

\begin{minipage}{0.48\textwidth}
\centering
\subfloat[SUPPORT ($N=8873$, $C=32\%$, $d=14$).\label{subtab:calibration_support}]{%
\begin{tabular}{lccc} 
{Model} & {ICI $\downarrow$} & {D-Cal $\uparrow$} & {C-Cal $\uparrow$} \\ \midrule
CoxPH \cite{cox_regression_1972} & 0.059 & \textbf{0.983}  & - \\
CoxNet \cite{simon_regularization_2011} & 0.059 & \textbf{0.983}  & - \\
CoxBoost \cite{hothorn_survival_2005} & \textbf{0.019} & 0.708  & - \\
RSF \cite{ishwaran_random_2008} & \textbf{0.022} & 0.532  & - \\
DSM \cite{nagpal_deep_2021} & 0.065 & 0.000  & - \\
DCM \cite{nagpal_deep_cox_2021} & 0.090 & \textbf{0.904}  & - \\
BNN-Cox (VI) \cite{qi_using_2023} & 0.073 & 0.096  & 0.952  \\
BNN-MTLR (VI) \cite{qi_using_2023} & \textbf{0.042} & 0.029  & 0.947  \\ \midrule
\textbf{\textit{This work:}}\\
Baseline (MLP) & 0.069 & 0.229  & - \\
+ VI & 0.064 & 0.000  & \textbf{0.977}  \\
+ MCD ($p=0.1$) & 0.077 & 0.000  & 0.938  \\
+ MCD ($p=0.2$) & 0.061 & 0.000  & \textbf{0.979}  \\
+ MCD ($p=0.5$) & 0.050 & 0.000  & \textbf{0.956}  \\
+ SNGP & 0.075 & 0.689  & - \\ \bottomrule
\end{tabular}}
\end{minipage}
\begin{minipage}{0.48\textwidth}
\centering
\subfloat[MIMIC-IV ($N=38520$, $C=67\%$, $d=91$).\label{subtab:calibration_mimic}]{%
\begin{tabular}{lccc} 
{Model} & {ICI $\downarrow$} & {D-Cal $\uparrow$} & {C-Cal $\uparrow$} \\ \midrule
CoxPH \cite{cox_regression_1972} & 0.037 & 0.549  & - \\
CoxNet \cite{simon_regularization_2011} & 0.033 & \textbf{0.553}  & - \\
CoxBoost \cite{hothorn_survival_2005} & \textbf{0.016} & \textbf{0.997}  & - \\
RSF \cite{ishwaran_random_2008} & 0.041 & \textbf{0.935}  & -\\
DSM \cite{nagpal_deep_2021} & \textbf{0.006} & 0.000  & - \\
DCM \cite{nagpal_deep_cox_2021} & 0.041 & 0.011  & - \\
BNN-Cox (VI) \cite{qi_using_2023} & 0.066 & 0.001  & 0.589  \\
BNN-MTLR (VI) \cite{qi_using_2023} & 0.057 & 0.060  & 0.583  \\ \midrule
\textbf{\textit{This work:}}\\
Baseline (MLP) & 0.026 & 0.068  & - \\
+ VI & 0.096 & 0.000  & 0.583  \\
+ MCD ($p=0.1$) & 0.034 & 0.014  & \textbf{0.600}  \\
+ MCD ($p=0.2$) & 0.038 & 0.013  & \textbf{0.596}  \\
+ MCD ($p=0.5$) & 0.036 & 0.008  & \textbf{0.597}  \\
+ SNGP & \textbf{0.015} & 0.230  & - \\ \bottomrule
\end{tabular}}
\end{minipage}
\end{table*}

Table \ref{tab:calibration_results} shows the calibration performance of our baseline MLP and its VI, MCD, and SNGP variants and the literature benchmarks in the test sets. ICI measures the mean absolute difference between the observed and predicted probabilities of an event. D-calibration expresses to what extent we can trust the predicted survival function, $S(t)$, according to a ${\chi}^2$ test. By dividing $S(t)$ into equally-sized intervals, it verifies whether the proportion of events in each interval is uniformly distributed. A $p$-value greater than 0.05 indicates that $S(t)$ is calibrated with respect to the event distribution. C-calibration expresses the agreement between the upper and lower bounds of the predicted CrIs and the observed intervals.

Regarding the baseline MLP model and its probabilistic variants, all models are D-calibrated on the small METABRIC dataset (see Table \ref{subtab:calibration_metabric}), with the VI and MCD variants also being C-calibrated. We assess the agreement between predicted and observed survival probabilities using the ICI metric, which is the mean absolute difference between the calibration curve and the diagonal line of perfect calibration. We see that the MCD models improve on the baseline in terms of ICI and thus provide a better-calibrated survival function. Using VI, despite being distribution and coverage calibrated, has a four-time higher ICI score than MCD in the METABRIC dataset. In the mid-sized SEER dataset (see Table \ref{subtab:calibration_seer}), all proposed architectures have D-calibration and C-calibration $p$-values greater than 0.05, indicating proper calibration; however, we observe a decrease in calibration performance in terms of ICI when using VI and MCD, also evidenced by lower $p$-values for D-calibration. In the SUPPORT dataset (see Table \ref{subtab:calibration_support}), only the baseline MLP and SNGP variants are statistically D-calibrated, and we see very poor D-calibration scores using VI or MCD. Indeed, increasing the complexity of the model (the number of parameters) and making the model more flexible can worsen D-calibration \cite{haider_effective_2020}.

We observe a similar trend in the large MIMIC-IV dataset (see Table \ref{subtab:calibration_mimic}), as none of the probabilistic methods are D-calibrated except for the SNGP. Regarding the tight coupling between the quality of the MCD uncertainty estimates and the parameter choices \cite{NIPS2016_8d8818c8}, e.g., the choice of dropout rate, we conduct empirical experiments with various dropout rates to measure its effect. In terms of calibration performance, we do not see significant differences using a low or high dropout rate; however, in the SEER, SUPPORT, and MIMIC-IV datasets, setting $p = 0.1$ provides higher $p$-values for D-calibration.

With respect to the existing models, our SNGP and MCD variants and the BNN-Cox model are among those that give the lowest ICI scores on average but do not necessarily provide D-calibrated survival functions. In the large MIMIC-IV dataset, neither the existing neural network models nor our proposed VI and MCD variants are D-calibrated, except for the BNN-MTLR model. In contrast, the SNGP approach, comprising a deterministic MLP with the output computed from a GP layer, is properly D-calibrated across all experimental datasets. It is important to rigorously assess multiple metrics to see how calibrated a model truly is because each calibration metric assesses a different kind of calibration, and it may be application-specific which type of calibration is needed. We note that the proposed models have been optimized according to the MLE objective in Eq. \eqref{eq:cox_ll}, which optimizes discriminative performance, not calibration performance.

\subsection{Model complexity}
\label{sec:model_complexcity}

Table \ref{tab:model_complexity} compares two literature benchmark models and our proposed models in terms of space complexity and number of trainable parameters on the MIMIC-IV dataset. For a fair comparison, we configure all models with a single hidden layer with 32 nodes. Notice the doubling of the number of parameters when using VI as opposed to the baseline MLP and MCD, resulting in increased overhead. Despite the fact that both are of $\mathcal{O}(d)$ complexity, doubling the number of trainable parameters can make a significant difference in high-dimensional models, so that VI is no longer an attractive solution.
\begin{table}[!htbp]
\centering
\caption{Space complexity of the models and number of trainable parameters (for the MIMIC-IV data, 91 covariates).}
\label{tab:model_complexity}
\begin{tabular}{@{}lcc@{}}
\toprule
\multicolumn{1}{c}{Model} & Space complex. & \#Params. \\ 
\midrule
BNN-Cox (VI) \cite{qi_using_2023} & $\mathcal{O}(d)$ & 5570 \\ 
BNN-MTLR (VI) \cite{qi_using_2023} & $\mathcal{O}(Kd)$ & 10718 \\
\midrule
\textbf{\textit{This work:}}\\
Baseline (MLP) & $\mathcal{O}(d)$ & 2849 \\
+ VI & $\mathcal{O}(d)$ & 5700 \\
+ MCD & $\mathcal{O}(d)$ & 2882 \\
+ SNGP & $\mathcal{O}(d)$ & 3776 \\ \bottomrule 
\end{tabular}
\end{table}

\begin{figure*}
\centering
\subfloat[Survival prediction using MCD ($p = 0.5$).]{%
\includegraphics[width=0.49\textwidth]{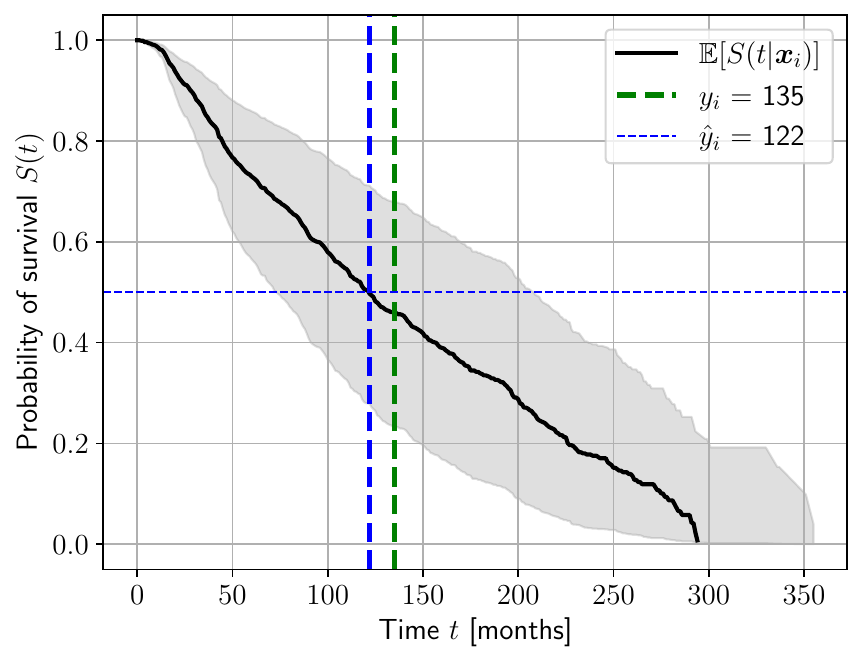}}
\subfloat[Histogram of $\hat{y}_{i}$ using MCD ($p = 0.5$).]{%
\includegraphics[width=0.505\textwidth]{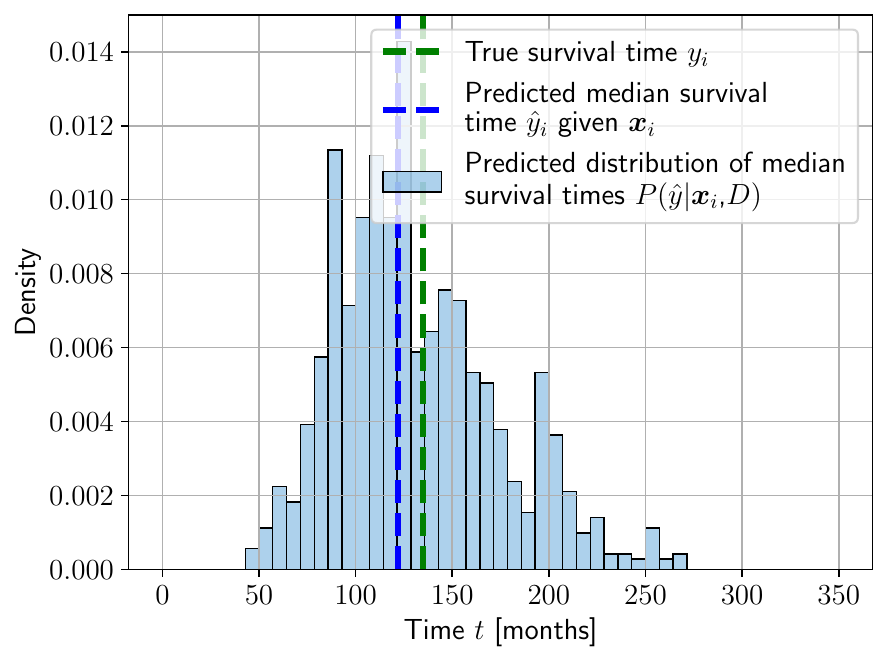}}
\\
\subfloat[Survival prediction using VI.]{%
\includegraphics[width=0.49\textwidth]{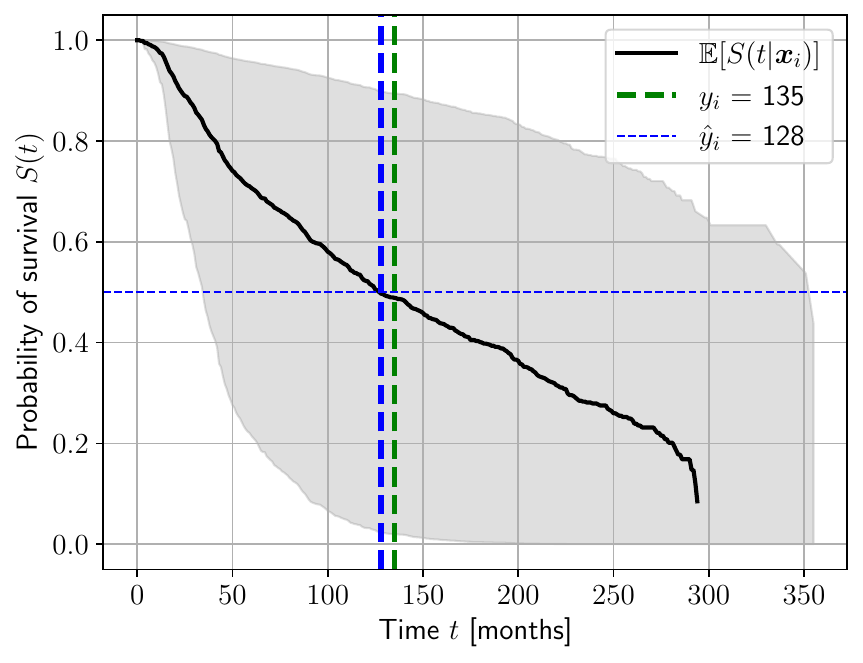}}
\subfloat[Histogram of $\hat{y}_{i}$ using VI.]{%
\includegraphics[width=0.505\textwidth]{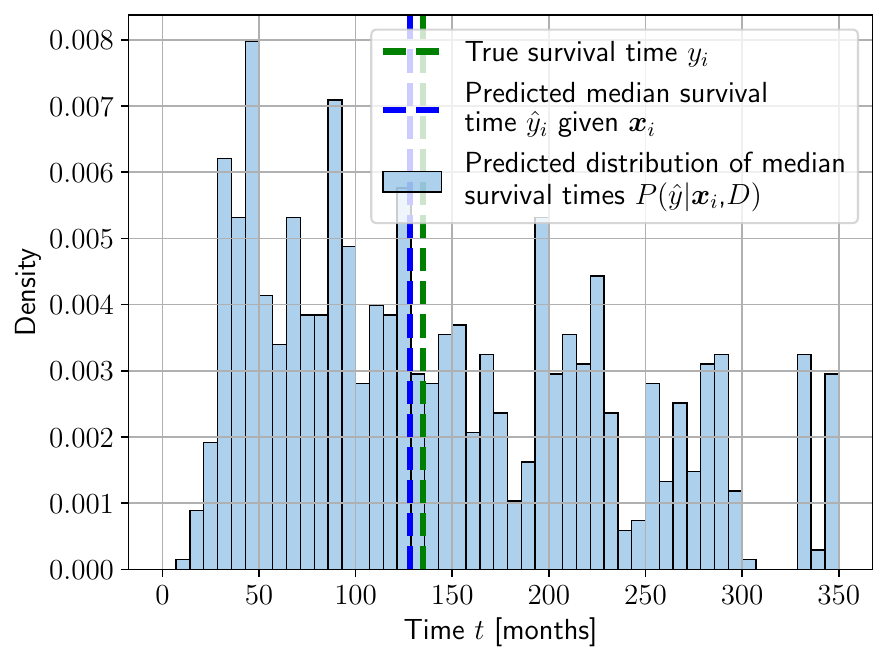}}
\caption{Model prediction on the METABRIC dataset using MCD and VI. For an individual $i$, we use the median of the predicted survival function as the predicted survival time $\hat{y}_i$. Although MCD and VI predict similar median survival times, VI has more variance around this particular prediction than MCD. Left column: mean individual survival function and its corresponding 90\% CrI. Right column: histogram (as an approximation of the underlying density) of the predicted survival time.}
\label{fig:model_predictioninference}
\end{figure*}

\subsection{Uncertainty estimation}
\label{sec:uncertainty_estimationmodel_inference}

We adopt the concept of Credible Intervals (CrIs) from \cite{qi_using_2023}. CrIs belong to the Bayesian framework and offer a way to quantify the uncertainty associated with a prediction. A credible interval is defined by two values, called the lower and upper bounds, which enclose a certain percentage of the predictive distribution. The main difference between CrIs and confidence intervals is that CrIs treat the parameter of interest as a random variable and the bounds as fixed values, while confidence intervals treat the parameter of interest as a fixed value and the bounds as random variables \cite[Ch. 4]{pml1Book}. Fig. \ref{fig:model_predictioninference} shows two ways in which uncertainty can be expressed along with a survival prediction for decision support purposes. For this, we selected a random sample from the METABRIC test set and drew 1000 samples from the predictive distribution to estimate its survival function. Although MCD and VI predict similar median survival times (in Fig. \ref{fig:model_predictioninference}, 122 and 128, respectively, where the actual time is 135), the latter predicts a greater amount of variance around the survival function (left column) and also the histogram of the predicted survival time displays higher variance (right column). The width of the CrIs expresses the uncertainty of the model, which is especially useful when making predictions about novel individuals (new data samples). This example underlines the importance of communicating the uncertainty associated with a prediction. Adopting a model that can express uncertainty about its predictions can help medical professionals make custom treatment plans with an additional level of risk assessment. This, in turn, can justify the level of trust and confidence that physicians and patients have in the prediction. Non-probabilistic models cannot offer these advantages.

\section{Conclusion}

In this work, we have compared three approaches for training probabilistic survival models for the first time in terms of prediction and calibration performance. The main finding of our work is that, evidently, MCD and SNGP constitute two attractive alternatives to VI for this task, as they rival VI's performance, but without the computational overhead. In many cases, our MCD model improved the mean absolute error over continuous state-of-the-art baselines while giving C-calibration scores similar to VI, and our SNGP model improved D-calibration over VI in all datasets. Limitations of our work include a lack of high-dimensional datasets (e.g., imaging or video data), where efficient computation is even more critical than in tabular data. Moreover, it is difficult for our models to simultaneously optimize prediction and calibration performance, as the results show. Future work should investigate how to improve the proposed model's predictions without degrading their calibration.

\section*{Conflict of interests}

The authors declare that they have no known conflicts of interest or personal relationships that could have appeared to influence the work reported in this paper.

\appendices

\section{Performance metrics}
\label{app:performance_metrics}

\textbf{CI\textsubscript{td}}: The time-dependent Concordance Index \cite{antolini_time-dependent_2005} is the probability that $\{i,j\}$ are concordant given they are comparable, where concordance is defined as such: given a comparable pair $\{i,j\}$, with event times $y_i < y_j$ and $\delta_i = 1$, the predicted survival probability for $j$ is greater than the predicted survival probability for $i$, at the time where $i$ experienced the event, and $j$ is still free from the event. This can be written as:

\begin{align}
\pi_{\text{comp}} &= \text{Pr}\bigl(y_i < y_j \; \& \; \delta_i = 1\bigr)\text{,}\\
\pi_{\text{conc}} &= \text{Pr}\bigl(S(y_i | \bm{X}_i(t)) < S(y_i | \bm{X}_j (t)) \nonumber \\
&\quad \; \& \; y_i < y_j ; \delta_i = 1\bigr)\text{,}\\
CI\textsubscript{td} &= \frac{\pi_{\text{conc}}}{\pi_{\text{comp}}} = \text{Pr}\bigl(S(y_i | \bm{X}_i(t)) < S(y_i | \bm{X}_j (t)) \nonumber \\
&\quad \; \& \; y_i < y_j ; \delta_i = 1\bigr)\text{,}
\end{align}

\noindent where $\pi_{\text{comp}}$ is the probability that $\{i,j\}$ are comparable, and $\pi_\text{conc}$ is the probability that $\{i,j\}$ are concordant.

\textbf{IBS}: The Integrated Brier Score \cite{graf_assessment_1999} is the expectation of the single-time Brier Score (BS) over some interval $\sbr{t_{k}, t_{K}}$,
where $t_{k}$ denotes some measurement time on some discretized event horizon of $K$ bins, i.e., $[t_0, t_1, t_2, \ldots, t_K]$. The BS is the mean squared error between the observed binary outcome $y_{i}(t_{k})$ and the predicted probability of the outcome $\hat{y_{i}}(t_{k})$ at $t_{k}$. If we consider the entire event horizon $\sbr{t_{0}, t_{K}}$, then the IBS is defined as \cite{graf_assessment_1999}:
\begin{equation}
IBS = \frac{1}{K} \sum_{t=0}^{K} \frac{1}{N} \sum_{i=1}^{N} (\hat{y}_{i}(t_{k}) - y_{i}(t_{k}))^2\text{.}
\end{equation}

\textbf{MAE:} The mean absolute error is the absolute difference between the predicted and actual survival times. We use the median of the survival function as the predicted survival time. Given an individual survival distribution, $S\br{t\;|\;\bm{x}_{i}} = \text{Pr}\br{T>t\;|\;\bm{x}_{i}}$, we calculate the predicted survival time $\hat{y}_i$ as the median survival time \cite{qi_survivaleval_2024}:
\begin{equation}\label{eq:yi}
\hat{y}_i = \text{median}\;(S(t\;|\;\bm{x}_{i})) = S^{-1}(0.5\;|\;\bm{x}_{i})\text{.}
\end{equation}
In this work, we report MAE\textsubscript{H} and MAE\textsubscript{PO}. The MAE\textsubscript{H} \cite{qi2023effective} uses hinge loss to calculate the mean absolute error for censored individuals: if the predicted survival time is smaller than the censoring time, the loss is the censoring time minus the predicted time. If the predicted survival time is equal to or greater than the censoring time, the loss is zero. MAE\textsubscript{PO} \cite{qi2023effective} employs pseudo-observation to estimate the actual time of survival for censored individuals.

\textbf{ICI:} The Integrated Calibration Index \cite{austin_ici_2019} is the absolute difference on average between the observed and predicted probabilities of the event. Let $f(x) = |x - x_{c}|$ denote the absolute difference between the calibration curve and the diagonal line of perfect calibration, where $x$ is a predicted probability and $x_{c}$ is the value of the calibration curve at $x$. Let $\phi(x)$ denote the density function of the predicted probabilities distribution. The ICI is then a weighted average of the absolute differences between the calibration curve and the diagonal line of perfect calibration, i.e., $ICI = \int_{0}^{1} f(x)\phi(x) \,dx$.

\textbf{D-calibration:} Distribution calibration \cite{haider_effective_2020} measures the calibration performance of $S(t)$, expressing to what extent the predicted probabilities can be trusted. For any probability interval $[a, b] \in [0, 1]$, we define $D_m(a, b)$ as the subset of individuals in the dataset $D$ whose predicted probability of event is in the interval $[a, b]$ \cite{qi2023effective}. A model is D-calibrated if the proportion of individuals $|D_m(a, b)|/|D|$ is statistically similar to the proportion $b-a$, which we assess using a ${\chi}^2$ test. For example, compare the time-of-death empirical quantiles against the computed ones: e.g., with a D-calibrated curve, approximately 50\% of these individuals should die prior to their common predicted median survival time.

\textbf{C-calibration}: Coverage calibration \cite{qi_using_2023} is a statistical test to measure the agreement between the upper and lower bounds of the predicted Credible Intervals (CrIs) and the observed probability intervals. For example, a 90\% CrI should cover an individual's likelihood of an event with a probability of 90\%. In practice, we calculate the observed coverage rates using CrIs with different percentages (ranging from 10\% to 90\%) and then use a ${\chi}^2$ test to evaluate the calibration of expected and observed coverage rates.

\section{Model hyperparameters}
\label{app:model_hyperparameters}

Table \ref{tab:model_hyperparameters} reports the selected hyperparameters. All proposed models use the same hyperparameters unless otherwise stated. We use Bayesian optimization \cite{snoek_practical_2012} to tune hyperparameters over ten iterations on the validation set, adopting the hyperparameters leading to the highest concordance index (CI\textsubscript{td}).

\begin{table}[!htbp]
 \caption{Selected hyperparameters for the different datasets.}
  \label{tab:model_hyperparameters} 
  \centering
  \begin{tabular}{lcccc}
    \toprule
    Hyperparameter & METABRIC & SEER & SUPPORT & MIMIC-IV  \\
    \cmidrule(lr){1-1} \cmidrule(lr){2-5}
    Optimizer & Adam & Adam & Adam & Adam  \\
    Decay & None & None & 0.001 & 0.0001 \\
    Activation fn & ReLU & ReLU & ReLU & ReLU \\
    Batch size & 32 & 32 & 32 & 128 \\
    \# Nodes per layer & [64, 128] & [64] & [128] & [32] \\
    Learning rate & $0.00001$ & $0.001$ & $0.001$ & $0.0001$ \\
    L2 regularization & 0.001 & 0.001 & 0.001 & 0.001 \\
    Dropout rate & 0.25 & 0.25 & 0.25 & 0.5 \\
    \bottomrule
  \end{tabular}
\end{table}

\section*{References}

\bibliographystyle{IEEEtran}
\bibliography{references}

\begin{thebibliography}{10}
\providecommand{\url}[1]{#1}
\csname url@samestyle\endcsname
\providecommand{\newblock}{\relax}
\providecommand{\bibinfo}[2]{#2}
\providecommand{\BIBentrySTDinterwordspacing}{\spaceskip=0pt\relax}
\providecommand{\BIBentryALTinterwordstretchfactor}{4}
\providecommand{\BIBentryALTinterwordspacing}{\spaceskip=\fontdimen2\font plus
\BIBentryALTinterwordstretchfactor\fontdimen3\font minus
  \fontdimen4\font\relax}
\providecommand{\BIBforeignlanguage}[2]{{%
\expandafter\ifx\csname l@#1\endcsname\relax
\typeout{** WARNING: IEEEtran.bst: No hyphenation pattern has been}%
\typeout{** loaded for the language `#1'. Using the pattern for}%
\typeout{** the default language instead.}%
\else
\language=\csname l@#1\endcsname
\fi
#2}}
\providecommand{\BIBdecl}{\relax}
\BIBdecl

\bibitem{gal_dropout_2016}
Y.~Gal and Z.~Ghahramani, ``Dropout as a {B}ayesian approximation: Representing
  model uncertainty in deep learning,'' in \emph{Proceedings of The 33rd
  International Conference on Machine Learning}, vol.~48, 2016, pp. 1050--1059.

\bibitem{magris_bayesian_2023}
M.~Magris and A.~Iosifidis, ``Bayesian learning for neural networks: an
  algorithmic survey,'' \emph{Artificial Intelligence Review}, vol.~56, no.~10,
  pp. 11\,773--11\,823, 2023.

\bibitem{katzman_deepsurv_2018}
J.~Katzman, U.~Shaham, J.~Bates, A.~Cloninger, T.~Jiang, and Y.~Kluger,
  ``{DeepSurv}: personalized treatment recommender system using a {Cox}
  proportional hazards deep neural network,'' \emph{BMC Medical Research
  Methodology}, vol.~18, no.~1, pp. 1--12, 2018.

\bibitem{nagpal_deep_2021}
C.~Nagpal, X.~Li, and A.~Dubrawski, ``Deep survival machines: Fully parametric
  survival regression and representation learning for censored data with
  competing risks,'' \emph{IEEE Journal of Biomedical and Health Informatics},
  vol.~25, no.~8, pp. 3163--3175, 2021.

\bibitem{nagpal_deep_cox_2021}
C.~Nagpal, S.~Yadlowsky, N.~Rostamzadeh, and K.~Heller, ``Deep {Cox} {Mixtures}
  for survival regression,'' in \emph{Machine Learning for Healthcare
  Conference}, 2021, pp. 674--708.

\bibitem{feng_bdnnsurv_2021}
D.~Feng and L.~Zhao, ``{BDNNSurv}: Bayesian deep neural networks for survival
  analysis using pseudo values,'' \emph{Journal of Data Science}, vol.~19,
  no.~4, pp. 542--554, 2021.

\bibitem{qi_using_2023}
S.-a. Qi, N.~Kumar, R.~Verma, J.-Y. Xu, G.~Shen-Tu, and R.~Greiner, ``Using
  {Bayesian} neural networks to select features and compute credible intervals
  for personalized survival prediction,'' \emph{IEEE Transactions on Biomedical
  Engineering}, vol.~70, no.~12, pp. 3389--3400, 2023.

\bibitem{loya2020uncertainty}
H.~Loya, P.~Poduval, D.~Anand, N.~Kumar, and A.~Sethi, ``Uncertainty estimation
  in cancer survival prediction,'' \emph{arXiv preprint arXiv:2003.08573},
  2020, accepted at AI4AH Workshop at ICLR 2020.

\bibitem{graves_practical_2011}
A.~Graves, ``Practical variational inference for neural networks,'' in
  \emph{Advances in Neural Information Processing Systems}, vol.~24, 2011, pp.
  2348--–2356.

\bibitem{lillelund_uncertainty_2023}
C.~M. Lillelund, M.~Magris, and C.~F. Pedersen, ``Uncertainty estimation in
  deep {B}ayesian survival models,'' in \emph{2023 IEEE EMBS International
  Conference on Biomedical and Health Informatics (BHI)}, 2023, pp. 1--4.

\bibitem{NIPS2011_1019c809}
C.-N. Yu, R.~Greiner, H.-C. Lin, and V.~Baracos, ``Learning patient-specific
  cancer survival distributions as a sequence of dependent regressors,'' in
  \emph{Advances in Neural Information Processing Systems}, vol.~24, 2011, pp.
  1845--1853.

\bibitem{gareth_introduction_2021}
J.~Gareth, W.~Daniela, H.~Trevor, and T.~Robert, \emph{An introduction to
  statistical learning: with applications in R}, 2nd~ed.\hskip 1em plus 0.5em
  minus 0.4em\relax Spinger, 2021.

\bibitem{cox_regression_1972}
D.~R. Cox, ``Regression models and life-tables,'' \emph{Journal of the Royal
  Statistical Society: Series B (Methodological)}, vol.~34, no.~2, pp.
  187--202, 1972.

\bibitem{breslow_analysis_1975}
N.~E. Breslow, ``Analysis of survival data under the proportional hazards
  model,'' \emph{International Statistical Review}, pp. 45--57, 1975.

\bibitem{gal2016uncertainty}
Y.~Gal, ``Uncertainty in deep learning,'' Ph.D. dissertation, University of
  Cambridge, 2016.

\bibitem{liu2020simple}
J.~Liu, Z.~Lin, S.~Padhy, D.~Tran, T.~Bedrax~Weiss, and B.~Lakshminarayanan,
  ``Simple and principled uncertainty estimation with deterministic deep
  learning via distance awareness,'' in \emph{Advances in Neural Information
  Processing Systems}, vol.~33, 2020, pp. 7498--7512.

\bibitem{NIPS2016_8d8818c8}
I.~Osband, C.~Blundell, A.~Pritzel, and B.~V. Roy, ``Deep exploration via
  bootstrapped {DQN},'' in \emph{Advances in Neural Information Processing
  Systems}, vol.~29, 2016, pp. 4033--4041.

\bibitem{metabric_group_genomic_2012}
{METABRIC Group}, C.~Curtis, S.~P. Shah, S.-F. Chin, G.~Turashvili, O.~M.
  Rueda, M.~J. Dunning, D.~Speed, A.~G. Lynch, S.~Samarajiwa, Y.~Yuan,
  S.~Gräf, G.~Ha, G.~Haffari, A.~Bashashati, R.~Russell, S.~McKinney,
  A.~Langerød, A.~Green, E.~Provenzano, G.~Wishart, S.~Pinder, P.~Watson,
  F.~Markowetz, L.~Murphy, I.~Ellis, A.~Purushotham, A.-L. Børresen-Dale,
  J.~D. Brenton, S.~Tavaré, C.~Caldas, and S.~Aparicio, ``The genomic and
  transcriptomic architecture of 2,000 breast tumours reveals novel
  subgroups,'' \emph{Nature}, vol. 486, no. 7403, pp. 346--352, 2012.

\bibitem{ries_cancer_2007}
L.~A. Gloeckler~Ries, M.~E. Reichman, D.~R. Lewis, B.~F. Hankey, and B.~K.
  Edwards, ``Cancer survival and incidence from the surveillance, epidemiology,
  and end results ({SEER}) program,'' \emph{Oncologist}, vol.~8, no.~6, pp.
  541--552, 2003.

\bibitem{knaus_support_1995}
W.~A. Knaus, ``The {SUPPORT} prognostic model: Objective estimates of survival
  for seriously ill hospitalized adults,'' \emph{Annals of Internal Medicine},
  vol. 122, no.~3, pp. 191--203, 1995.

\bibitem{johnson_mimic-iv_2023}
A.~E.~W. Johnson, L.~Bulgarelli, L.~Shen, A.~Gayles, A.~Shammout, S.~Horng,
  T.~J. Pollard, S.~Hao, B.~Moody, B.~Gow, L.-w.~H. Lehman, L.~A. Celi, and
  R.~G. Mark, ``{MIMIC}-{IV}, a freely accessible electronic health record
  dataset,'' \emph{Scientific Data}, vol.~10, no.~1, pp. 1--9, 2023.

\bibitem{simon_regularization_2011}
N.~Simon, J.~Friedman, T.~Hastie, and R.~Tibshirani, ``Regularization paths for
  {C}ox's proportional hazards model via coordinate descent,'' \emph{Journal of
  Statistical Software}, vol.~39, no.~5, pp. 1--13, 2011.

\bibitem{friedman_greedy_2001}
J.~H. Friedman, ``{Greedy function approximation: A gradient boosting
  machine.}'' \emph{The Annals of Statistics}, vol.~29, no.~5, pp. 1189--1232,
  2001.

\bibitem{ishwaran_random_2008}
H.~Ishwaran, U.~B. Kogalur, E.~H. Blackstone, and M.~S. Lauer, ``Random
  survival forests,'' \emph{The Annals of Applied Statistics}, vol.~2, no.~3,
  pp. 841--860, 2008.

\bibitem{snoek_practical_2012}
J.~Snoek, H.~Larochelle, and R.~P. Adams, ``Practical bayesian optimization of
  machine learning algorithms,'' in \emph{Advances in Neural Information
  Processing Systems}, vol.~25, 2012, pp. 2951--–2959.

\bibitem{antolini_time-dependent_2005}
L.~Antolini, P.~Boracchi, and E.~Biganzoli, ``A time-dependent discrimination
  index for survival data,'' \emph{Statistics in Medicine}, vol.~24, no.~24,
  pp. 3927--3944, 2005.

\bibitem{qi_survivaleval_2024}
S.-a. Qi, W.~Sun, and R.~Greiner, ``{SurvivalEVAL}: A comprehensive open-source
  {P}ython package for evaluating individual survival distributions,'' in
  \emph{Proceedings of the 2023 AAAI Fall Symposia}, vol.~2, no.~1, 2023, pp.
  453--457.

\bibitem{graf_assessment_1999}
E.~Graf, C.~Schmoor, W.~Sauerbrei, and M.~Schumacher, ``Assessment and
  comparison of prognostic classification schemes for survival data,''
  \emph{Statistics in Medicine}, vol.~18, no. 17-18, pp. 2529--2545, 1999.

\bibitem{austin_ici_2019}
P.~C. Austin and E.~W. Steyerberg, ``The integrated calibration index {(ICI)}
  and related metrics for quantifying the calibration of logistic regression
  models,'' \emph{Statistics in Medicine}, vol.~38, no.~21, pp. 4051--4065,
  2019.

\bibitem{haider_effective_2020}
H.~Haider, B.~Hoehn, S.~Davis, and R.~Greiner, ``Effective ways to build and
  evaluate individual survival distributions,'' \emph{Journal of Machine
  Learning Research}, vol.~21, no.~1, pp. 1--63, 2020.

\bibitem{hothorn_survival_2005}
T.~Hothorn, P.~Bühlmann, S.~Dudoit, A.~Molinaro, and M.~J. Van Der~Laan,
  ``{Survival ensembles},'' \emph{Biostatistics}, vol.~7, no.~3, pp. 355--373,
  2005.

\bibitem{pml1Book}
K.~P. Murphy, \emph{Probabilistic Machine Learning: An introduction}.\hskip 1em
  plus 0.5em minus 0.4em\relax MIT Press, 2022.

\bibitem{qi2023effective}
S.-a. Qi, N.~Kumar, M.~Farrokh, W.~Sun, L.~Kuan, R.~Ranganath, R.~Henao, and
  R.~Greiner, ``An effective meaningful way to evaluate survival models,'' in
  \emph{Proceedings of the 40th International Conference on Machine Learning},
  vol. 202, 2023, pp. 28\,244--28\,276.

\end{thebibliography}

\end{document}